%% file: main.tex
\documentclass{article}
\usepackage{spconf,graphicx}
\usepackage{amsmath}
\usepackage{amsfonts}
\usepackage[noend]{algpseudocode}
\usepackage{booktabs}
\usepackage{algorithm}      % For the floating 'algorithm' environment
\usepackage{algpseudocode}  % For the algorithmic styling
\usepackage{url}
\usepackage[hidelinks]{hyperref}
\usepackage{multirow}
\usepackage{pifont}
\usepackage{ifthen}
\graphicspath{ {images/} }
\usepackage{url}
\usepackage{hyperref}
\usepackage{microtype}
\usepackage[table]{xcolor}
\usepackage[
   backend=biber,
   style=ieee,
   citestyle=numeric-comp,
   maxbibnames=3,
   maxcitenames=3,
   doi=false,isbn=false,url=false,eprint=false
]{biblatex}
\input{sourcemap_definitions.tex}
\defbibheading{bibliography}[\refname]{}

\newboolean{showcomments}
\setboolean{showcomments}{true} % toggle to show or hide comments
\ifthenelse{\boolean{showcomments}}
  {\newcommand{\nb}[3]{
    \textcolor{#2}{\small [#3 --#1]}
   }
  }
  {\newcommand{\nb}[3]{}
  }

\DeclareMathOperator*{\argmax}{arg\,max}
\title{Enhancing End-to-End Conversational Speech Translation Through Target Language Context Utilization}
\name{
\begin{tabular}{c}
Amir Hussein$^1$, Brian Yan$^2$, Antonios Anastasopoulos$^3$,  \\Shinji Watanabe$^2$,
Sanjeev Khudanpur$^1$ 
\end{tabular}
}
\address{$^1$Johns Hopkins University, $^2$Carnegie Mellon University, $^3$George Mason University\\USA}
\begin{document}
\ninept
\maketitle
\begin{abstract}
Incorporating longer context has been shown to benefit machine translation, but the inclusion of context in end-to-end speech translation (E2E-ST) remains under-studied. To bridge this gap, we introduce target language context in E2E-ST, enhancing coherence and overcoming memory constraints of extended audio segments. Additionally, we propose context dropout to ensure robustness to the absence of context, and further improve performance by adding speaker information. Our proposed contextual E2E-ST outperforms the isolated utterance-based E2E-ST approach. Lastly, we demonstrate that in conversational speech, contextual information primarily contributes to capturing context style, as well as resolving anaphora and named entities.  
\end{abstract}
\begin{keywords}
Speech translation, contextual information, end-to-end models, conversational speech
\end{keywords}
\section{Introduction}
\label{sec:intro}
% motivation
Speech translation (ST) is crucial for breaking language barriers and enhancing global communication. Seamlessly translating spoken conversations across languages  impacts cross-cultural interactions, education, and diplomacy \cite{koksal2020role}. 
% Traditionally, ST systems have been built by cascading automatic speech recognition (ASR) and machine translation (MT) models . However, recently end-to-end speech translation (E2E-ST) approaches, in which the source speech is directly translated to the target text, are attracting more attention \cite{yan2023espnet}.
Traditionally, ST systems have been built by cascading automatic speech recognition (ASR) and machine translation (MT) models \cite{stentiford1990machine, ansari2020findings,waibel1991janus,bertoldi2005new,sperber2017toward,pino2019harnessing,yang2022jhu}. However, recently end-to-end speech translation (E2E-ST) approaches, in which the source speech is directly translated to the target language text, have gained more attention \cite{berard2016listen, berard2018endtoend, dalmia2021searchable, gaido2020end, yan2023espnet}. 
Despite the promising results of recent research advancements in E2E-ST systems, the produced translations often lack consistency in translation. 
% Despite the promising results, E2E-ST systems lack contextual consistency in translations.

Consistency is crucial for language understanding, as the meaning of utterances often depends on the broader conversational context. In MT, source-side context is commonly utilized to address inaccurate choice of pronouns \cite{guillou2016incorporating}, mistranslations of ambiguous words \cite{gonzales2017improving}, and general incoherence in translation \cite{voita2019good}. Several MT studies showed that document level context \cite{wang2017exploiting,miculicich2018document, voita2018context,lopes2020document,zheng2021towards} outperforms sentence level context \cite{10094687,hori2021advanced, kim2019acoustic,kim2019cross,radford2022robust,tian2017make,gain2021not}. Analogous to MT, incorporating contextual information in E2E-ST systems is expected to be valuable for coherent translation, resolving anaphora, and disambiguating words, especially homophones. To study the effectiveness of context on E2E-ST, researchers have used simple concatenation of audio input as context and the corresponding translation as target output, reporting improvements in pronoun and homophone translation~\cite{zhang2021beyond}. However, utilizing source-side context comes with the challenge of encoding very long audio segments, which can easily lead to memory bottlenecks, especially with self-attention based networks \cite{vaswani2017attention}.

To address this limitation, we propose a context-aware E2E-ST that leverages context in the output (target) side. In particular, in a conversational context, we incorporate the previously output sentences (in the target language) as the initial condition on the decoder side to generate the translation of the current input utterance. This enables the decoder to effectively utilize textual information from longer utterances and focus on vital parts of the previous context for more accurate and consistent translation. 
Unlike existing work, we focus on \textit{conversational} speech translation, an essential facet of daily communication which presents unique challenges: 1) high context dependence for meaning, 2) informal and grammatically inconsistent language usage, and 3) data scarcity. Our study covers three language pairs: Tunisian Arabic-English, Spanish-English, and Chinese-English. Our contributions encompass: (i) a context-aware E2E-ST framework employing target language context, (ii) enhanced robustness to context absence through context dropout, and (iii) context enrichment with speaker information. 
As an additional contribution, we conduct an ablation study to assess the significance of each component (context size, context dropout, and speaker information) on final performance. Finally, we perform a part-of-speech-based analysis to identify where the primary improvements result from the incorporation of context.
 
\section{CONTEXTUAL E2E-ST}
\label{sec:proposed-approach}
In standard E2E-ST, the 
% task is formulated as Bayesian decision. The 
goal is to find the most probable target word sequence $\hat{\mathbf{Y}}$ of length $N$, out of all possible outputs $\mathbf{Y}^*$. This is done by selecting the sequence that maximizes the posterior likelihood $P(\mathbf{Y}|\mathbf{X})$, given a $T$-long sequence of $D$-dimensional speech features, represented as $\mathbf{X} = \{\mathbf{x}_t \in \mathbb{R}^D| t = 1, \cdots, T \}$. In our approach we incorporate $K$ \textit{previous} translations $\mathbf{Y}^{\mathrm{cntx}}=\{\mathbf{y}_l \in \mathbb{V}^{\mathrm{tgt}}| l = 1, \cdots, K \}$ in the target language $\mathbb{V}^{\mathrm{tgt}}$. We incorporate the context as an initial condition on the decoder side. Therefore, our objective is to maximize the posterior likelihood given both the input speech and the context:
\begin{equation}
    %\begin{aligned}
    \hat{\mathbf{Y}} = %& \argmax_{Y^*} log(P(Y|X)) \\ 
    %& = 
    \argmax_{\mathbf{Y}^*} \sum_{l=1}^{N}\log(P(\mathbf{Y}_l|\mathbf{Y}_{<l}, \mathbf{X}, \mathbf{Y}^{\mathrm{cntx}}))
    %\end{aligned}
\end{equation}
We enrich the context with speaker role information, encoded as speaker tags within $\mathbb{V}^{\mathrm{tgt}}$. Figure~\ref{ST-diagram} presents an illustrative example showcasing a target sentence along with a context of its two preceding sentences. This context is augmented with speaker role information: [SpkA] and [SpkB] represent to the first and second speaker in the conversation, respectively. The [SEP] tag indicates separation between the sentences within the context. 
\begin{quote}
    \begin{align*}
        [\textbf{Context}] & \text{\ [SpkA] I'm from Peru, and you? [SEP]} \\
         & \text{\ [SpkB] Puerto Rico.} \\
         [\textbf{Target}] & \text{\ [SpkA] Oh, from Puerto Rico, oh, ok.}
    \end{align*}
    %\textit{ {[\textbf{Context}]} [SpkA] I'm from Peru, and you? [SEP] [SpkB] Puerto Rico. [\textbf{Target}] [SpkA] Oh, from Puerto Rico, oh, ok.}
\end{quote}
It's important to note that, unlike \cite{kanda2020serialized} where speaker labels are used for prediction, we employ them solely in the context as initial condition and do not predict them.

\subsection{E2E-ST Architecture}
\begin{figure}[t]
\centering
 % \vspace{-.8cm}
%\includegraphics[width=6.2cm,height=6.5cm]{images/ST-ctc-attention.pdf}
\includegraphics[width=.85\columnwidth]{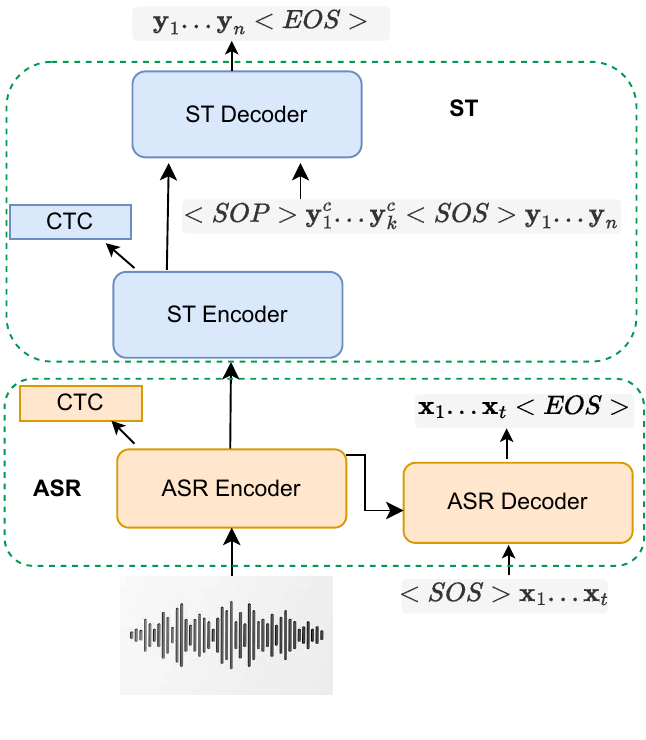}
\vspace{-1em}
\caption{Illustration of proposed contextual E2E-ST approach.}
\label{ST-diagram} 
 \vspace{-1em}
\end{figure}

Our proposed contextual E2E-ST builds upon the CTC/Attention architecture composed of conformer encoders with hierarchical CTC encoding and transformer decoders~\cite{yan2023ctc, yan2023espnet} as depicted in Fig~\ref{ST-diagram}. 
The CTC/Attention approach decomposes ST into ASR and MT encoder-decoder models. The ASR encoder, $\mathrm{ENC_{asr}}(\cdot)$, maps input speech $\mathbf{X}$ into hidden representation $\mathrm{H^E_{asr}}$ shown in Eq. (\ref{eq1}).
\begin{equation}
    H^\mathrm{E}_{\mathrm{asr}} = ENC_{\mathrm{asr}}(\mathbf{X})
    \label{eq1}
\end{equation}
Following this, the representation $\mathrm{H^E_{asr}}$ from Eq. (\ref{eq1}) serves as input to both the ST encoder, $\mathrm{ENC_{st}}(\cdot)$, and the ASR decoder, $\mathrm{DEC_{asr}}(\cdot)$, as demonstrated in Eq. (\ref{eq2},\ref{eq5}) 
\begin{equation}
    H^\mathrm{E}_{\mathrm{st}} = ENC_{\mathrm{st}}(H^E_{\mathrm{asr}})
    \label{eq2}
\end{equation}
\begin{equation}
    H^\mathrm{D}_{\mathrm{asr}} = DEC_{\mathrm{asr}}(H^E_{\mathrm{asr}})
    \label{eq5}
\end{equation}
Given $\mathrm{H^E_{st}}$ from Eq. (\ref{eq2}), the ST encoder  $\mathrm{DEC_{st}}(\cdot)$ generates $\hat{\mathbf{y}}_l$ at each time step that is conditionally dependent on the hidden representation $\mathrm{H^E_{st}}$, previous target sequence $\mathbf{Y}_{1:t-1}$ and the context $\mathbf{Y}^{\mathrm{cntx}}$ of previous sentences shown in Eq. (\ref{eq3}, \ref{eq4}). 
\begin{equation}
    H^\mathrm{D}_{\mathrm{st}} = DEC_{\mathrm{st}}(H^E_{\mathrm{st}}, \mathbf{Y}^{\mathrm{cntx}}_{1:K}, \mathbf{Y}_{1:l-1})
    \label{eq3}
\end{equation}
\begin{equation}
    P(\mathbf{y}_l|\mathbf{Y}_{1:l-1},\mathbf{X}, \mathbf{Y}^{\mathrm{cntx}}) = \text{Softmax}(H^\mathrm{D}_{\mathrm{st}})
    \label{eq4}
\end{equation}
% \begin{align*}
%     &H^\mathrm{E}_{\mathrm{asr}} = ENC_{\mathrm{asr}}(\mathbf{X}) \\
%     &H^\mathrm{E}_{\mathrm{st}} = ENC_{\mathrm{st}}(H_{\mathrm{asr}}) \\
%     &H^\mathrm{D}_{\mathrm{st}} = DEC_{\mathrm{st}}(H_{\mathrm{st}}, \mathbf{Y}^c_{1:K}, \mathbf{Y}_{1:l-1}) \\
%     &P(\mathbf{y}_l|\mathbf{Y}_{1:l-1},\mathbf{X}, \mathbf{Y}^c) = Softmax(H^\mathrm{D}_{\mathrm{st}})
% \end{align*}
The unique characteristic of our approach lies in the use of context also during training. While other methods may attempt to predict the context, our model uses the context exclusively as an initial condition for the decoder, overcoming memory constraints of extended audio. During model training, the ST cross-entropy loss is computed solely for the target translation, excluding the context. The model is optimized with a multi-task learning objective combining hybrid ASR attention $\mathcal{L}^{\mathrm{asr}}_{\mathrm{att}}$ and CTC $\mathcal{L}^{\mathrm{asr}}_{\mathrm{ctc}}$ as well as hybrid ST attention $\mathcal{L}^{\mathrm{st}}_{\mathrm{att}}$ and CTC $\mathcal{L}^{\mathrm{st}}_{\mathrm{ctc}}$ losses:

\begin{equation}\label{eq:6}
    \begin{aligned}
    \mathcal{L} = &\alpha_3((1-\alpha_1)\mathcal{L}^{\mathrm{asr}}_{\mathrm{att}} + \alpha_1\mathcal{L}^{\mathrm{asr}}_{\mathrm{ctc}})\\
    & + (1-\alpha_3)((1-\alpha_2)\mathcal{L}^{\mathrm{st}}_{\mathrm{att}} + \alpha_2\mathcal{L}^{\mathrm{st}}_{\mathrm{ctc}})
    \end{aligned}
\end{equation}
where $\alpha's$ are used for interpolation.

\section{EXPERIMENTS}
\label{sec:pagestyle}
\subsection{Dataset}
% data description
We demonstrate the efficacy of our proposed approach through evaluations on three conversational datasets (their statistics are summarized in Table~\ref{tab:datasets}): Fisher-CallHome Spanish English \cite{post2013improved}, IWSLT22 Tunisian Arabic-English \cite{antonios2022findings}, and BOLT Chinese-English \cite{song2014collecting}. 
% combined with extra translations obtained manually by BBN.\footnote{\url{https://www.bbntranslations.com/}} 
These datasets contain $3$-way data comprising telephone speech, source language transcriptions, and corresponding English translations.   
\begin{table}[hbt!]
    \centering
     %\vspace{-0.2cm}
     \vspace{-1em}
     \caption{Statistics for the conversational ST corpora.}
    \include{tables/datasets}
    \vspace{-1em}
    
    \label{tab:datasets}
     %\vspace{-1em}
\end{table}
% describe the any pre/ post processing of the data
 We use separated source and target vocabularies, each consisting of $4$K byte-pair-encoding \cite{kudo2018sentencepiece} (BPE) units. All audios are resampled from $8$kHz to $16$kHz, augmented with speed perturbations ($0.9$, $1.0$ and $1.1$) and transformed into $83$-dimensional feature frames ($80$ log-mel filterbank coefficients plus $3$ pitch features). Additionally, we augment the features with \texttt{specaugment}~\cite{park2019specaugment}, with mask parameters $(mT,mF,T,F)=(5,2,27,0.05)$ and bi-cubic time-warping. During scoring we report the results in terms of case-sensitive BLEU with punctuation. We also measure the statistical significance of improvement using paired bootstrap resampling with sacreBLEU\cite{post-2018-call}.

\subsection{Baseline configuration}
We conduct all experiments by customizing the ESPnet toolkit \cite{yan2023espnet}. For the encoders $\mathrm{ENC_{asr}}$ in Eq. (\ref{eq1}) and $\mathrm{ENC_{st}}$ in Eq. (\ref{eq2}), we employ the conformer architecture \cite{gulati2020conformer} consisting of 12 blocks for $\mathrm{ENC_{asr}}$ and $6$ blocks for $\mathrm{ENC_{st}}$. Both encoders are configured with $2048$ feed-forward dimensions, $256$ attention dimensions, and $4$ attention heads. The transformer architecture is employed for $\mathrm{DEC_{asr}}$ and $\mathrm{DEC_{st}}$ in Eq. (\ref{eq5}), each with $6$ decoder blocks and the same configuration as the encoders. The model has $72$M parameters. We follow ST best practices~\cite{yan2023espnet}, we first pretrain the ASR module followed by fine-tuning of the entire E2E-ST model for the translation task. Our training configuration remains consistent for both pretraining and fine-tuning, employing Adam optimizer with a learning rate of $0.001$, warmup-steps of $25$K, a dropout-rate of $0.1$ and $40$ epochs. We use joint training with hybrid CTC/attention by setting CTC weight ($\alpha_1$, $\alpha_2$) in Eq. (\ref{eq:6}) to $0.3$ and the weight that combines ASR and ST losses ($\alpha_3$) to $0.3$. During inference, we use a beam size of $10$ and length penalty of $0.3$.

\subsection{Contextual E2E-ST configuration}
% training: explain the context (gold)
In the Contextual E2E-ST model (described in \S\ref{sec:proposed-approach}), we retain the same configuration as the baseline. The only difference is that during training with teacher-forcing, the decoder initial condition is the preceding sentences alongside a start-of-sentence token. Heuristically, we limit any long contextual sentences to the last 50 tokens. Upon visual inspection, we found that these truncated sentences effectively capture the contextually relevant information. The previous sentence considered a part of context only if they are from the same recording.\footnote{The first utterance of each conversation has no associated context.}
We will refer to (Gold) context when employing ground-truth translations and to (Hyp) when utilizing the model's predictions. During training we only use the Gold context, however during inference we explore both Gold and Hyp context, simulating both an oracle and a more realistic scenario.

\section{Results}
\subsection{Contextual ST results}
% gold previous vs random gold context
We examined the effect of preceding context on model performance by comparing Gold previous context with the no-context baseline and randomly selected sentences unrelated to the ground truth. The results, shown in Table~\ref{tab:gold-rnd}, use a context size of one preceding sentence.   
 \begin{table}[tb!]
  % \vspace{-0.2cm}
    \centering
    \vspace{-1em}
    \caption{Comparison of the BLEU scores according to contextual information quality using one previous utterrance as context. $\dagger$: denotes a statistically significant difference ($p < 0.01$) compared to the no-context baseline.}
    \include{tables/gold-vs-rnd-context}

    \label{tab:gold-rnd}
     \vspace{-1em}
\end{table}
% The analysis reveals that when compared to the no-context baseline, using gold context consistently yields improvements with maximum BLEU increase of +1.5. 
% On the other hand, the random context consistently degrades performance, to a maximum BLEU score reduction of -0.7. 
The analysis indicates that, compared to the no-context baseline, gold context consistently improves performance, with a maximum BLEU increase of $+1.5$. Random context, on the other hand, consistently lowers performance up to a maximum BLEU reduction of $-0.7$.  This outcome affirms that incorporating even a single sentence as context yields improvements, which stem exactly from high-quality context rather than other artifacts.

\subsection{Context dropout}
\label{subsec:cx-dropout}
% reduce bias with context dropout
Next we explore the bias resulting from training with gold context when no context is available during inference (results in Table~\ref{tab:cx-dropout}). 
A model trained with context but applied to inference without context (third row) shows a degradation of almost $-1$ BLEU points even compared to the non-contextual baseline (first row). 
This clearly demonstrates the context-trained model's strong inclination to depend on context during inference. 
To overcome this limitation we propose to use context dropout: during training, 
target-side context is probabilistically included or not. We experiment with various percentages of context dropout ([$0.2$--$0.7$]) and find that $0.2$ yields the best results across all datasets. 
Now, using context dropout (row four) leads to an improvement of up to $+0.6$ in BLEU score when context is available during inference. 
But more importantly, row five shows that the  model trained with context dropout exhibits robustness to inference without context, even slightly surpassing the context-less baseline's performance.
\begin{table}[t]
    \centering
     %\vspace{-0.2cm}
     \vspace{-1em}
    \caption{Comparison of the BLEU scores according to the bias towards contextual information using context size of one. $\dagger$ denotes a statistically significant difference ($p < 0.01$) compared to the no-context baseline.}
    \include{tables/context-dropout}

    \label{tab:cx-dropout}
\end{table}

\subsection{Context size and speaker role}
\label{subsec:cx-spk}

\begin{figure}[b!]
\centering
\vspace{-1em}
\includegraphics[width=8.8cm,height=5cm]{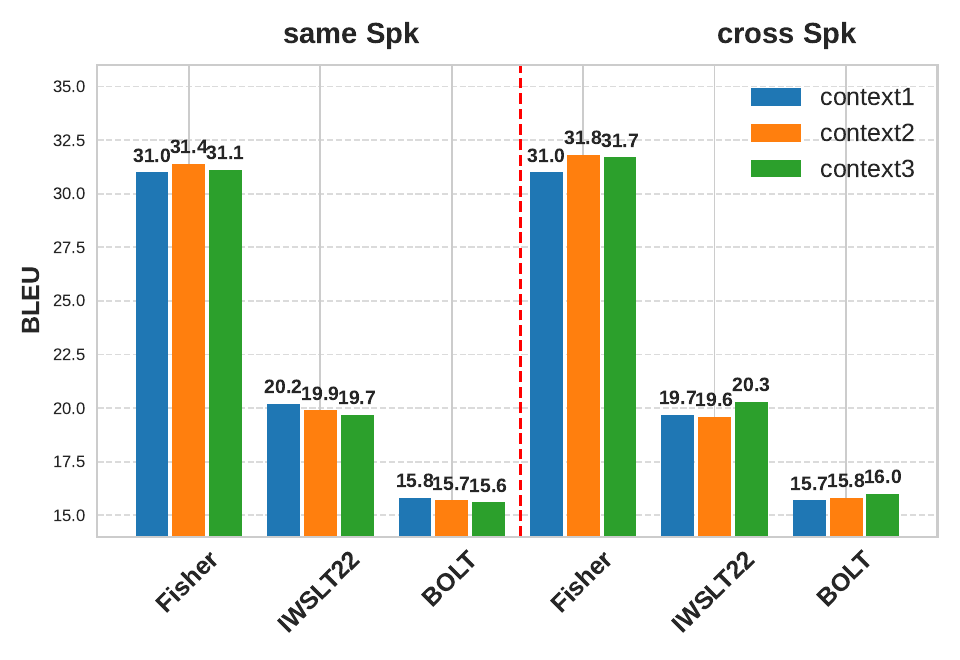}
\vspace{-1em}
\caption{Comparison of the BLEU scores according to context size and speaker information. Cross-speaker context outperforms same-speaker context, but the optimal context size varies (2--3).}
\label{fig:cx-size} 
\vspace{-1em}
\end{figure}

In this part we investigate the impact of context size and speaker information. We explore previous gold context of size [1,2,3], which may include utterances from a different speaker (cross Spk) or we can select context only from the same speaker. Results are in Fig~\ref{fig:cx-size}.
% Below is an illustrative example showcasing a target sentence along with a context of size two, enriched with cross-speaker information using distinct tags that encapsulate conversational speaker identities: 
% \begin{quote}
%     \textit{ {[\textbf{Context}]} [SpkA] I'm from Peru, and you? [SEP] [SpkB] Puerto Rico. [\textbf{Target}] [SpkA] Oh, from Puerto Rico, oh, ok.}
% \end{quote}
From Fig \ref{fig:cx-size}, we conclude that cross-speaker context consistently outperforms same-speaker context, evident in the BLEU scores showing improvements of $+.4$, $+0.2$, and $+0.1$ for Fisher, BOLT, and IWSLT22 respectively. The optimal context size for cross speaker is between $2$--$3$: $2$ for Fisher, $3$ for IWSLT22 and BOLT. % \an{what about how much context? (1,2,3) Needs a sentence on this}

\begin{figure*}[bt!]

    \centering
   \includegraphics[width=17.5cm,height=5cm]{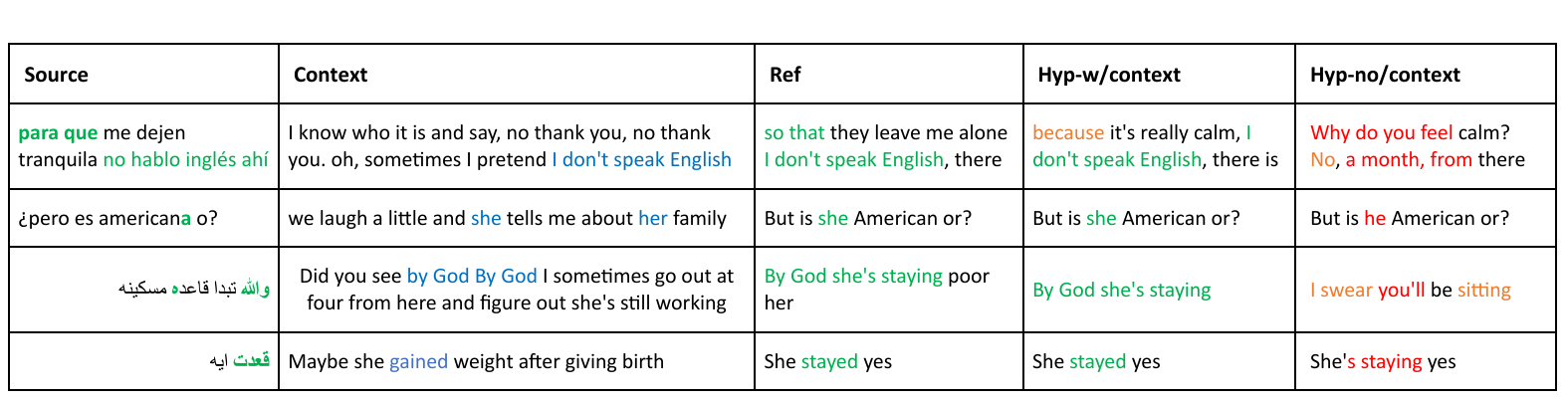}
    \vspace{-1em}
    \caption{Examples of translations with and without context. Color code: \{\textbf{Green}: matching with reference; \textbf{Red}: incorrect; \textbf{Orange}: not matching with reference but similar meaning; \textbf{Blue}: related contextual tokens\} 
    % \an{Convert table to LaTeX native instead of using image.}
    }
    \label{fig:cx-examples}
    \vspace{-1em}
    %\vspace{-3.0cm}
\end{figure*}

\subsection{Using model-predicted context}
Now we turn to using model-predicted context (Hyp) instead of Gold context, using the optimal configuration from \S(\ref{subsec:cx-dropout},\ref{subsec:cx-spk}) with $0.2$ context dropout and cross-speaker context. We examine two decoding approaches: a) \textbf{Exact decoding:} at every step, the model's previous predictions are used as the contextual input for subsequent predictions. This method may exhibit error propagation because initial predictions can affect later ones. b) \textbf{Multi-stage decoding:} initial predictions from isolated utterances form the context for subsequent predictions. This method controls context dependence, by adjusting the number of decoding stages and enables parallelization, as the initial predictions can be made independently. Additionally, we provide a comparative analysis against the best results achieved using gold context, shown in Table~\ref{tab:cx-hyp-gold}. The overall improvement of using Gold context, compared to the baseline, is up to $+2.2$ BLEU points.
 \begin{table}[bt!]
    \centering
    \vspace{-1em}
    \caption{Comparison of using predicted Hyp context (\textbf{Exact} and \textbf{Multistage}), and best Gold context results. \textbf{Context Size} column indicates the number of previous utterances used as context for each respective dataset. $\dagger$: denotes a statistically significant difference ($p < 0.01$) compared to the no-context baseline.}
    \include{tables/gold-hyp-context}

    \label{tab:cx-hyp-gold}
    \vspace{-2em}
\end{table}
On the other hand, it is noteworthy that Hyp Multistage context outperforms Hyp Exact. For multistage decoding, we use only one stage as we noticed a degradation after the first stage, which supports our error propagation hypothesis. The overall improvement using Multistage approach compared to the no-context baseline is up to $+0.9$ BLEU. Additionally, on the BOLT dataset, the Multistage approach nearly matches the results achieved with Gold context. 
% still need to compare to model prediction exact and multistage

\subsection{Where do we improve?}
\begin{figure}[bt!]
\centering
\includegraphics[width=.98\columnwidth]{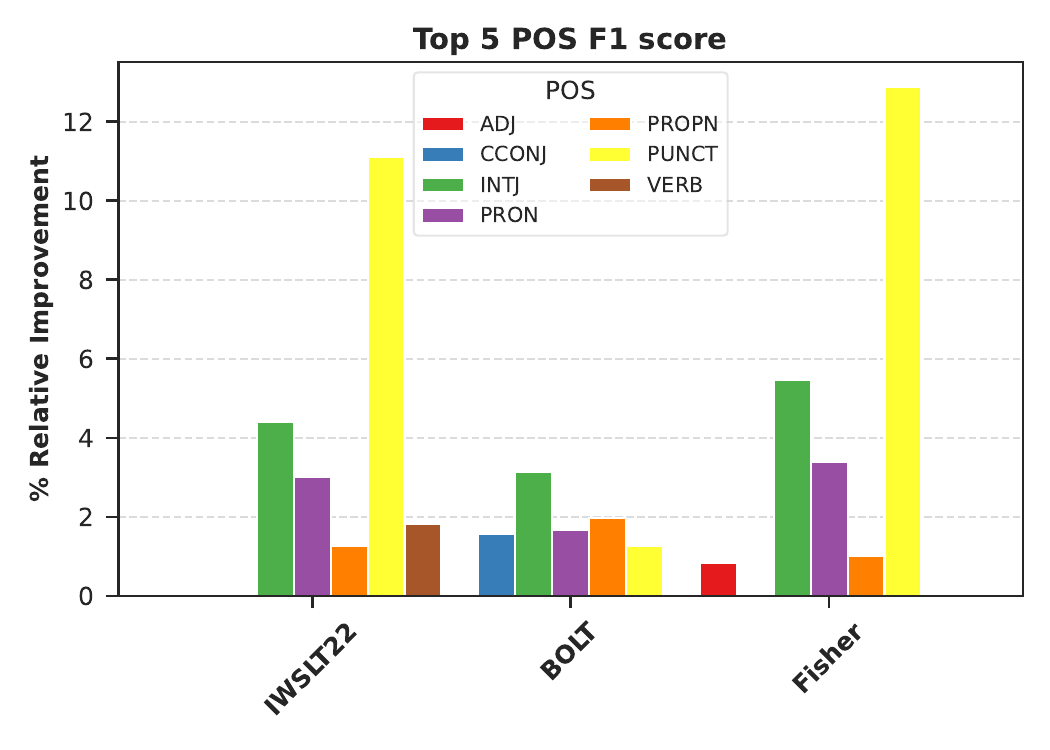}
\vspace{-1em}
\caption{Top 5 relative improvements in F1 score across different Part-of-Speech Tags.}
\label{fig:pos} 
\vspace{-2em}
%\vspace{-3.0cm}
\end{figure}

% BLEU scores, being an aggregate over the whole test corpus, may fail to capture the nuances that a context-informed model may capture.
While BLEU scores provide an aggregate measure over the entire test corpus, they may fail to capture the nuances that a context-informed model can address. Even if the difference in BLEU scores is minor, our model retains critical contextual information. To better understand where our model improves, we analyze the part-of-speech (POS) tags of the predictions compared to the reference, using a state-of-the-art transformer-based POS-tagger from Spacy.\footnote{\url{https://spacy.io/models/en}} We focus on Spanish (Fisher), IWSLT22, and BOLT. We compute the relative improvement in F1 score for each POS tag and visualize the top five improvements in Fig~\ref{fig:pos}.
We observe a consistent pattern where the highest relative improvement is from \textit{Context Style} (PUNCT, INTJ) followed by \textit{Anaphora} (PRON), and \textit{Entities} (PROPN). It is worth noting that a large portion of BOLT dataset is not punctuated, which contributes to the relatively lower PUNCT improvement.
Examples highlighting the benefits of contextual information are presented in Figure~\ref{fig:cx-examples}. 
In these examples, context helps in disambiguating: a) homophones: \textit{para que  $\rightarrow$ so that} vs \textit{por que $\rightarrow$ why}, b) pronouns: \textit{americana o} vs \textit{americano} $\rightarrow$ \textit{she} vs \textit{he is American} 
c) coherence in style: \textit{By God} vs \textit{swear} d) verb tense: \textit{stayed} vs \textit{staying}.  

\section{Conclusion}
We developed an end-to-end contextual Speech Translation (ST) model that leverages target language context, significantly 
% \an{Did we run statistical significance?} 
outperforming a no-context baseline across three conversational speech translation datasets. This highlights the pivotal role of high-quality contextual information, including cross-speaker information, in enhancing model performance. Despite context-trained models exhibiting strong contextual dependence during inference, this can be effectively mitigated by implementing context dropout. Moreover, the comparison between different decoding strategies, using model-predicted context, showed that a multi-stage decoding approach provides significant improvements and reduces the risk of error propagation. Our analysis demonstrates that contextual information contributes primarily to context style, anaphora, and entities.

\section{ACKNOWLEDGEMENTS}
\label{sec:print}
This work was carried out during the 2023 Summer Camp for Applied Language Exploration at Human Language Technology Center of Excellence. In addition, this work was partially supported by NSF CCRI Grant No 2120435.

% Below is an example of how to insert images. Delete the ``\vspace'' line,
% uncomment the preceding line ``\centerline...'' and replace ``imageX.ps''
% with a suitable PostScript file name.
% -------------------------------------------------------------------------
% \begin{figure}[htb]

% \begin{minipage}[b]{1.0\linewidth}
%   \centering
%   \centerline{\includegraphics[width=8.5cm]{image1}}
% %  \vspace{2.0cm}
%   \centerline{(a) Result 1}\medskip
% \end{minipage}
% %
% \begin{minipage}[b]{.48\linewidth}
%   \centering
%   \centerline{\includegraphics[width=4.0cm]{image3}}
% %  \vspace{1.5cm}
%   \centerline{(b) Results 3}\medskip
% \end{minipage}
% \hfill
% \begin{minipage}[b]{0.48\linewidth}
%   \centering
%   \centerline{\includegraphics[width=4.0cm]{image4}}
% %  \vspace{1.5cm}
%   \centerline{(c) Result 4}\medskip
% \end{minipage}
% %
% \caption{Example of placing a figure with experimental results.}
% \label{fig:res}
% %
% \end{figure}

%\vfill
%\pagebreak

% \section{COPYRIGHT FORMS}
% \label{sec:copyright}

% You must submit your fully completed, signed IEEE electronic copyright release
% form when you submit your paper. We {\bf must} have this form before your paper
% can be published in the proceedings.

\vfill\pagebreak

% \bibliographystyle{IEEEbib}
% \bibliography{strings,refs}
\section{REFERENCES}
\label{sec:refs}
\printbibliography

\end{document}

%% file: sourcemap_definitions.tex
\DeclareSourcemap{
  \maps[datatype=bibtex, overwrite=true]{
    \map{
      \step[fieldsource=booktitle, match=\regexp{.*Interspeech.*}, replace={Proc. Interspeech}]
      \step[fieldsource=journal, match=\regexp{.*INTERSPEECH.*}, replace={Proc. Interspeech}]
      \step[fieldsource=booktitle, match=\regexp{.*ICASSP.*}, replace={Proc. ICASSP}]
      \step[fieldsource=booktitle, match=\regexp{.*icassp_inpress.*}, replace={Proc. ICASSP (in press)}]
      \step[fieldsource=booktitle, match=\regexp{.*Acoustics,.*Speech.*and.*Signal.*Processing.*}, replace={Proc. ICASSP}]
      \step[fieldsource=booktitle, match=\regexp{.*International.*Conference.*on.*Learning.*Representations.*}, replace={Proc. ICLR}]
      \step[fieldsource=booktitle, match=\regexp{.*International.*Conference.*on.*Computational.*Linguistics.*}, replace={Proc. COLING}]
      \step[fieldsource=booktitle, match=\regexp{.*SIGdial.*Meeting.*on.*Discourse.*and.*Dialogue.*}, replace={Proc. SIGDIAL}]
      \step[fieldsource=booktitle, match=\regexp{.*International.*Conference.*on.*Machine.*Learning.*}, replace={Proc. ICML}]
      \step[fieldsource=booktitle, match=\regexp{.*North.*American.*Chapter.*of.*the.*Association.*for.*Computational.*Linguistics:.*Human.*Language.*Technologies.*}, replace={Proc. NAACL}]
      \step[fieldsource=booktitle, match=\regexp{.*Empirical.*Methods.*in.*Natural.*Language.*Processing.*}, replace={Proc. EMNLP}]
      \step[fieldsource=booktitle, match=\regexp{.*Association.*for.*Computational.*Linguistics.*}, replace={Proc. ACL}]
      \step[fieldsource=booktitle, match=\regexp{.*Automatic.*Speech.*Recognition.*and.*Understanding.*}, replace={Proc. ASRU}]
      \step[fieldsource=booktitle, match=\regexp{.*Spoken.*Language.*Technology.*}, replace={Proc. SLT}]
      \step[fieldsource=booktitle, match=\regexp{.*Speech.*Synthesis.*Workshop.*}, replace={Proc. SSW}]
      \step[fieldsource=booktitle, match=\regexp{.*workshop.*on.*speech.*synthesis.*}, replace={Proc. SSW}]
      \step[fieldsource=booktitle, match=\regexp{.*Advances.*in.*neural.*information.*processing.*}, replace={Proc. NeurIPS}]
      \step[fieldsource=booktitle, match=\regexp{.*Advances.*in.*Neural.*Information.*Processing.*}, replace={Proc. NeurIPS}]
      \step[fieldsource=booktitle, match=\regexp{.*Workshop.*on.*Applications.*of.*Signal.*Processing.*to.*Audio.*and.*Acoustics.*}, replace={Proc. WASPAA}]
      \step[fieldsource=publisher, match=\regexp{.+}, replace={{}}]
      \step[fieldsource=month, match=\regexp{.+}, replace={{}}]
      \step[fieldsource=location, match=\regexp{.+}, replace={{}}]
      \step[fieldsource=address, match=\regexp{.+}, replace={{}}]
      \step[fieldsource=organization, match=\regexp{.+}, replace={{}}]
    }
  }
}

%% file: tables/datasets.tex
\resizebox{0.45\textwidth}{!}{
\begin{tabular}{l|c|c|c|c}
\toprule
 \multirow{2}{*}{\textbf{Corpus}}& \multirow{2}{*}{\textbf{Lang}} & \multicolumn{3}{c}{\textbf{\#Hours}}\\ 
 \cmidrule{3-5} 
& & \textbf{Train} & \textbf{Dev} & \textbf{Test} \\ \midrule
Spanish Fisher/Callhome & Sp-En & 186.3 & 9.3 & 4.5/1.8 \\ 
Tunisian IWSLT22 & Ar-En & 161.0 &  6.3 & 3.6 \\ 
Chinese BOLT      & Zh-En & 110.6  & 8.5 &  8.5 \\
\bottomrule
\end{tabular}}

%% file: tables/gold-vs-rnd-context.tex
% \resizebox{0.48\textwidth}{!}{
% \begin{tabular}{l cc cc cc}
% \toprule
% %\multirow{2}{*}{\textbf{Context}} & \multicolumn{4}{c|}{\textbf{Sp-En}}  & \multicolumn{2}{c}{\textbf{Ar-En}} \\ \cmidrule{2-7}
%  %& \multicolumn{4}{c|}{\textbf{Sp-En}}  & \multicolumn{2}{c}{\textbf{Ar-En}} \\ \cmidrule{2-7}
% & \multicolumn{2}{c}{\textbf{Fisher}} & \multicolumn{2}{c}{\textbf{CallHome}}  & \multicolumn{2}{c}{\textbf{IWSLT22}} \\
% \cmidrule{2-5} \cmidrule{6-7} 

% & BLEU & ChrF2 & BLEU & ChrF2 &BLEU & ChrF2\\ \midrule
% no context & & & & & & \\
% \bottomrule
% \end{tabular}}

% \resizebox{0.48\textwidth}{!}{
% \begin{tabular}{l cc cc cc}
% \toprule
% & \multicolumn{2}{c}{\textbf{Fisher}} & \multicolumn{2}{c}{\textbf{CallHome}}  & \multicolumn{2}{c}{\textbf{IWSLT22}} \\
% \cmidrule(lr){2-3} \cmidrule(lr){4-5} \cmidrule(lr){6-7}
% Context Type & BLEU & BLEU-lc & BLEU & BLEU-lc & BLEU & BLEU-lc \\ \midrule
% No context &29.8 & 34.4 & 25.9 & 27 &19.7 &22.2 \\
% Rand & 29.9& 34.1 & 25.2 & 26.5 &19.5 & 41.2 \\
% Gold & \textbf{31.3}& \textbf{34.6} & \textbf{26} &\textbf{27} & \textbf{19.9}&\textbf{22.2} \\
% \bottomrule
% \end{tabular}}

\resizebox{0.38\textwidth}{!}{
\begin{tabular}{l c c c}
\toprule
\textbf{Context Type} & \textbf{Fisher} & \textbf{CallHome}  & \textbf{IWSLT22} \\ \midrule
% \cmidrule(lr){2-3} \cmidrule(lr){4-4}
% Context Type & BLEU  & BLEU  & BLEU  \\ \midrule
No-context & 29.8  & 25.9 &19.7  \\
Random & 29.9 & 25.2 & 19.5  \\
Gold & \textbf{31.3}$\dagger$& \textbf{26.0} & \textbf{19.9} \\
\bottomrule
\end{tabular}}

%% file: tables/context-dropout.tex
\resizebox{0.48\textwidth}{!}{
\begin{tabular}{c c c c | c c c}
\toprule
 & \textbf{Train}& \textbf{Decode}& \textbf{Context}&\textbf{Fisher} & \textbf{CallHome}  & \textbf{IWSLT22} \\
\textbf{ID} & \textbf{w/context}&\textbf{w/context}& \textbf{Dropout} & & & \\ \midrule
\texttt{1} & \ding{55} & \ding{55} & - & 29.8 & 25.9 & 19.7 \\
\texttt{2} & \ding{51} & \ding{51} & - & \textbf{31.3}$\dagger$ & 26.0 & 19.9 \\
\texttt{3} & \ding{51} & \ding{55} & - & 29.3 & 25.0 & 18.9 \\
\texttt{4} & \ding{51} & \ding{51} & 0.2 & 31.0$\dagger$ & \textbf{26.5}$\dagger$ & \textbf{20.2}$\dagger$\\
\texttt{5} & \ding{51} & \ding{55} & 0.2 & 30.1 & 25.8& 19.8 \\
\bottomrule
\end{tabular}}

%% file: tables/gold-hyp-context.tex
\resizebox{0.48\textwidth}{!}{
\begin{tabular}{l l |c c c c}
\toprule
\textbf{Context}&\textbf{Context}& \multicolumn{3}{c}{\textbf{Evaluation Sets}} \\
  % Adjusted to span columns 3 through 5 only
\textbf{type}&\textbf{size} & \textbf{Fisher} & \textbf{CallHome} & \textbf{IWSLT22} &\textbf{BOLT} \\ 
\midrule
Baseline & no-context & 29.8 & 25.9 & 19.7 & 15.5 \\
Gold & (2,2,3,3) & \textbf{31.8}$\dagger$ & \textbf{28.1}$\dagger$ & \textbf{20.3}$\dagger$ &  \textbf{16.0}$\dagger$\\
\midrule
Hyp Exact & (2,2,3,3) & 30.2$\dagger$ & 25.9 & 19.8 & 15.6 \\
Hyp Multistage & (2,2,3,3) & \textbf{30.7}$\dagger$& \textbf{26.4}$\dagger$ & \textbf{19.8} & \textbf{15.9}$\dagger$\\
\bottomrule
\end{tabular}
}